\documentclass[journal,transmag]{IEEEtran}

\usepackage[english]{babel}
\usepackage[utf8x]{inputenc}
\usepackage[T1]{fontenc}

\usepackage{amsmath}
\usepackage{amssymb}
\usepackage{graphicx}
\usepackage{subcaption}
\usepackage{algorithm}
\usepackage{algpseudocode}
\usepackage[colorlinks=true, allcolors=blue]{hyperref}

\begin{document}

\title{Magnetic Field Prediction Using Generative Adversarial Networks}

\author{\IEEEauthorblockN{Stefan Pollok, Nataniel Olden-Jørgensen, Peter Stanley Jørgensen, and Rasmus Bjørk}
\vspace{1ex}
\IEEEauthorblockA{\small Department of Energy Conversion and Storage, Technical University of Denmark, 2800 Kgs. Lyngby, Denmark}
\thanks{Corresponding author: S. Pollok (e-mail: spol@dtu.dk).}}

\markboth{\textsc{Pollok} \MakeLowercase{\textit{et al.}}: Magnetic Field Prediction Using Generative Adversarial Networks}
{}

\IEEEtitleabstractindextext{%
\begin{abstract}
Plenty of scientific and real-world applications are built on magnetic fields and their characteristics. To retrieve the valuable magnetic field information in high resolution, extensive field measurements are required, which are either time-consuming to conduct or even not feasible due to physical constraints. To alleviate this problem, we predict magnetic field values at a random point in space from a few point measurements by using a generative adversarial network (GAN) structure. The deep learning (DL) architecture consists of two neural networks: a generator, which predicts missing field values of a given magnetic field, and a critic, which is trained to calculate the statistical distance between real and generated magnetic field distributions. By minimizing this statistical distance, a reconstruction loss as well as physical losses, our trained generator has learned to predict the missing field values with a median reconstruction test error of 5.14\%, when a single coherent region of field points is missing, and 5.86\%, when only a few point measurements in space are available and the field measurements around are predicted. We verify the results on an experimentally validated field.
\end{abstract}
}

\maketitle
\IEEEdisplaynontitleabstractindextext
\IEEEpeerreviewmaketitle

\section{Introduction}
\label{sec:introduction}
Magnetic fields are used in a multitude of scientific and real-world applications, from MRI scanners to electric motors. In all of these applications the magnetic field must be optimized for the given technology, and the magnetic field in each device must typically be characterized for this. However, to characterize a magnetic field, this must be determined throughout the volume of interest, regardless of whether the magnetic field is measured using a Hall sensor in an experimental setup or the field is computed using a simulation framework such as analytical modeling \cite{bjoerk21} or finite element analysis \cite{comsol20}. Determining the magnetic field with increasing resolution is computationally expensive, as is measuring the field in a large number of points for characterizing the field of an experimental setup. The problem of obtaining a detailed magnetic field from a set of measurement or simulation points in known in a number of fields.

In robotics, Gaussian processes have been used to interpolate and extrapolate magnetic field values from a few given data points \cite{solin18}. As the computational complexity of Gaussian processes renders the approach more or less useless in practice, when the number of observations becomes large, i.e., more than several thousand measurements, the authors model a scalar potential function instead of the 3-D magnetic field. In addition to that, the presented method uses an approximation of the covariance function to model the ambient magnetic field, which inherits information of magnetic field disturbances from the surrounding indoor environment. A robot is performing localization using this information and the following robot navigation is highly dependent on the quality of the magnetic field estimation. Le Grand et al. \cite{grand12} use a simple linear interpolation of the measured mesh points to perform a mapping from coarse, expensive magnetic field measurements to a fine magnetic field estimate, which inherently is a low-order approximation.

In magnetohydrodynamics, the dynamics of conducting fluids have to be described. The predictions of the charged fluid particle trajectories rely on the exact magnetic field values in each location. Given numerical results of a magnetic field evaluation on a discrete grid, the divergence-free magnetic field values at a random point in space are obtained by relating the magnetic field to its vector potential using Fourier transforms \cite{mackay06}. The resultant vector potential is then interpolated using cubic splines. Bernauer et al. \cite{bernauer16} use a spline-based interpolation.

In geophysics, least-squares collocation is used for the interpolation of the earth anomaly map from given magnetic measurements of different sources \cite{maus09}. Another approach to model the geomagnetic field on the Earth's surface is to interpolate the external magnetic field disturbances by using Spherical Elementary Current Systems \cite{mclay10}.

Moreover, there exist problems, where magnetic field values simply cannot be obtained and have to be interpolated. For instance, the photospheric magnetic field in the Sun’s polar region is unavailable in specific locations and in order to infer large-scale characteristics, the missing field data is interpolated \cite{sun11}. Here, the estimation of missing field data is performed with a third-order 2-D polynomial functions fitted to the given data by using least squares.

As demonstrated above the interpolations methods used for magnetic field differ between fields, as does the numerical accuracy and computational cost of the implementation. Here, we present a novel technique for interpolating and extrapolating magnetic field values based on deep learning (DL). DL is a data-driven approach that can be trained in a supervised manner and is proven to be an universal approximator \cite{hornik89}. Recent advances in this research field have led to outstanding results in natural language processing \cite{devlin19}, speech recognition \cite{amodei16}, and computer vision \cite{krizhevsky14}. The technique has not only proven to be beneficial for the mentioned engineering tasks, DL architectures are now used to solve challenges in natural sciences, e.g., material discovery \cite{bombarelli18} or drug design \cite{jumper21}.

In magnetism research, neural networks have been used, e.g., for approximately solving Maxwell’s equations for electromagnetic structures \cite{badrinarayanan15} or for solving Maxwell’s equations in an inverse manner \cite{pollok21}, i.e., inferring the magnetic structure from a given magnetic field. Recently, physics-informed neural networks (PINNs) \cite{raissi19} have been formulated to embed the nonlinear partial differential equations of a physical domain, into the DL architecture. That setup makes the prior knowledge of the problem's physics available to the DL method and therefore respects the given constraints during training. An instance of PINNs was adapted to the area of magnetism \cite{kovacs21}, where Maxwell's equations describe the underlying physical laws of magnetostatics and micromagnetics.

The underlying physical laws have been embedded in a recently emerging DL architecture called generative adversarial networks (GANs) \cite{goodfellow14}. In that setup, two neural networks are trained: a generator, which outputs a desired target sample, and a discriminator, which checks whether an output sample is real or artificially generated. By adding loss terms to the generator, which relate the generated target samples to the underlying stochastic differential equations, stochastic processes can be approximated \cite{yang18}. GANs have been used to generate images based on conditions \cite{mirza14}. Song et al. \cite{song21} learn a diffusion process from data to noise and by approximating the reverse-time stochastic differential equation, an image can be retrieved from only a few given parts of the original image. Another promising work \cite{oord16} learns to predict a probability distribution for each pixel value based on its preceding pixel neighbours. Further developments in their architecture and learning procedure have led to the ability to fill in missing pixel values of an RGB image in order to create a visually appealing and consistent output \cite{yu18}. Zheng et al. \cite{zheng20} enhance existing work for semantic image inpainting by loss terms, which introduce the physical constraints of a geostatistical problem, i.e., to infer the heterogeneous geological field of a few point measurements.

\begin{figure}[t]
\centering
\includegraphics[scale=0.38]{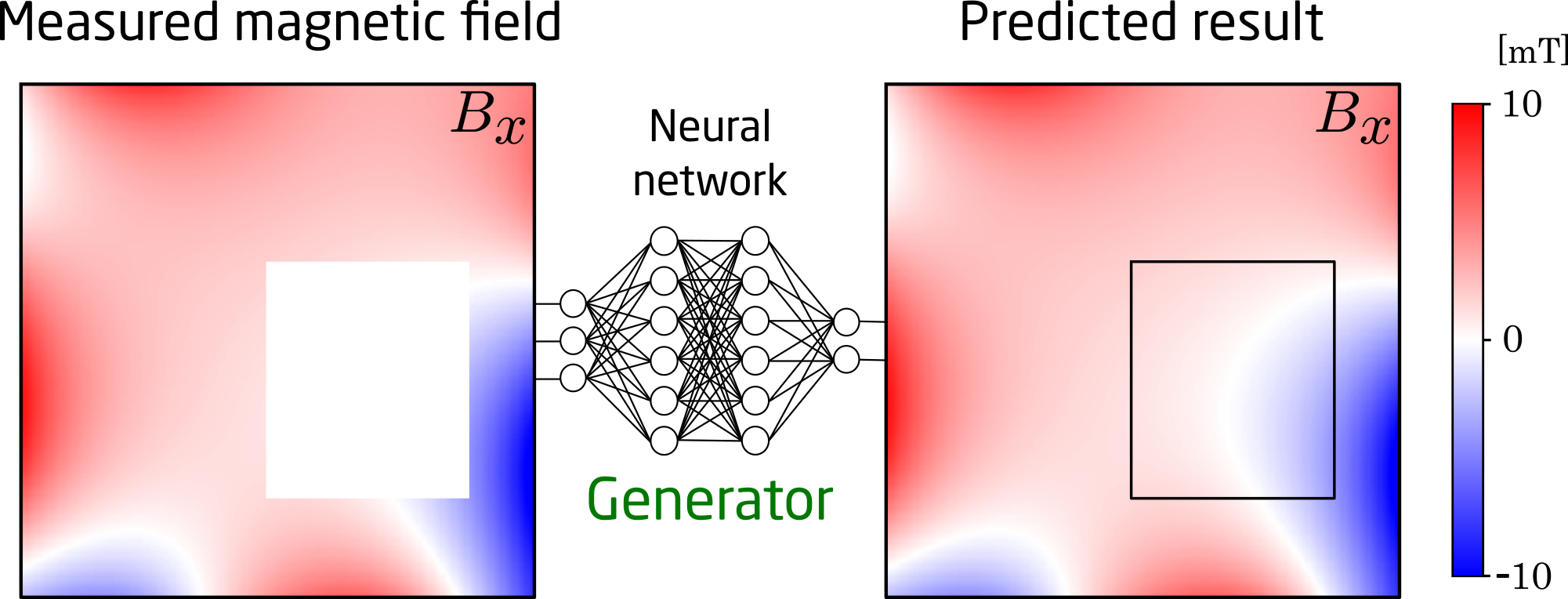}
\caption{Simplified overview of novel DL approach for magnetic field prediction. A generator neural network is trained to predict the missing magnetic field values.}
\label{fig:task}
\end{figure}

Here, we present a novel approach, where a physics-informed GAN is used to predict missing field values of a magnetic field. Whereas in principle the inpainted region of an image can be of any color as long as it is appealing for the human eye, the distribution and the behavior of a magnetic field are governed by Maxwell’s equations. By embedding the physical constraints into the loss function of our DL method, we show that the quality of our predicted field regions can be improved. A generator neural network $G$ shall reconstruct the real, underlying magnetic field $B$ by inter- and extrapolation on sparsely measured field values $B_{measured}$:

\begin{equation}
    B = G( B_{measured} ).
\end{equation}

Based on the partly measured field values, we consider two distinct problems. One of that is inpainting, where the the magnetic field is given around an area of unknown field values, which are then interpolated by our method as shown in Fig. \ref{fig:task}. The second task, which we call outpainting, is a combination of inter- and extrapolation. Hereby, magnetic field values are sparsely measured and the trained neural network is to generate the missing values.

To the best of our knowledge, this is the first application of GANs to magnetic field prediction. We extend previous work \cite{yu18} to an outpainting task and embed the physical behaviour of magnetic fields into additional losses, which the generator neural network is trained on. In addition to the performance of our novel method, we provide an extensive comparison to other state-of-the-art methods used for magnetic field prediction in literature and also compare to magnetic fields measured in a physical experiment. Hereby, we measure a magnetic field produced by multiple hard magnets with a Hall sensor and predict missing field values with our trained generator network.

\begin{figure*}[ht]
\centering
\includegraphics[scale=0.25]{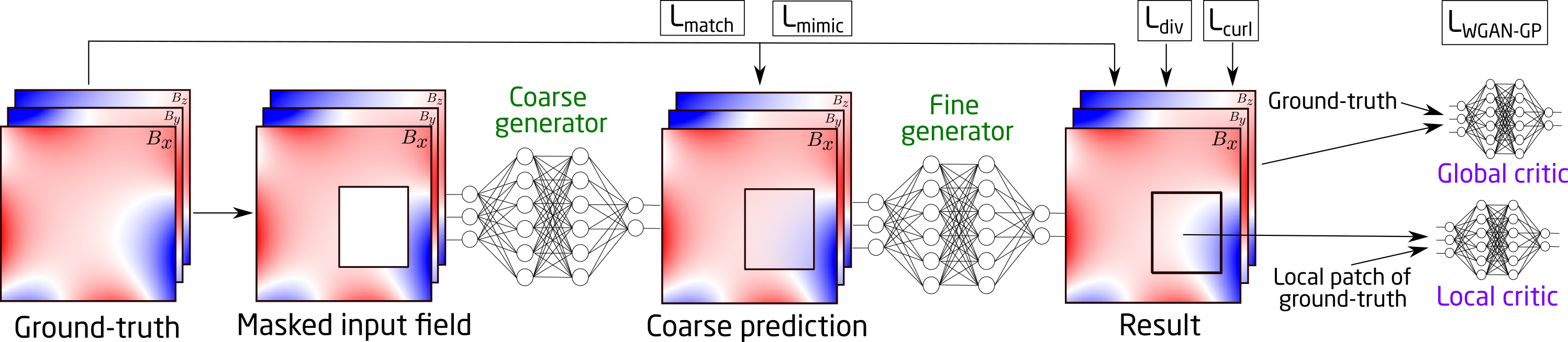}
\caption{Illustration of the DL architecture used for magnetic field prediction. A two-step generation process, which consists of down- and upsampling across multiple convolutional layers, produces missing field values of a masked input magnetic field. The result is evaluated by a local and a global critic, which are again neural networks consisting of several convolutional layers. In addition to the $L_{WGAN-GP}$, the $l1$ reconstruction losses, $L_{match}$ and $L_{mimic}$, and the physical losses, $L_{div}$ and $L_{curl}$, are calculated for updating the parameters of the generator networks in order to minimize the overall loss function. }
\label{fig:dl_architecture}
\end{figure*}

\section{Physics-informed GANs}
A method capable of addressing the introduced problem is physics-informed GAN. By supervision, a generator neural network $G$ learns to produce samples that match the distribution of the ground-truth training data, which are magnetic fields in our case. Given some measured magnetic field values in this area as input, the trained $G$ outputs a complete magnetic field from the learned distribution, which is constrained to match the given measurements and to meet physical properties of magnetic fields.

\subsection{Wasserstein GAN with Gradient Penalty}
In the original formulation of GANs, two neural networks are competing in a game. A generator network $G$ maps a sample $\mathbf{z}$ of a simple noise distribution to a sample $\mathbf{x}$ of the model distribution $\mathbb{P}_g$ as $G(z;\theta_g)$, where $\theta_g$ are the trainable network parameters. Simultaneously, a discriminator network $D$ is trained to output a scalar for a given sample in the form $D(\mathbf{x};\theta_d)$, where $\theta_d$ are its network parameters to be optimized. The idea is that $G$ tries to fool $D$ by generating samples that resemble the ones taken from the real target space distribution $\mathbb{P}_r$. On the other hand, $D$ improves in distinguishing real from generated samples during training. That should force $G$ to generate even more realistic samples by minimizing the Jensen-Shannon divergence between $\mathbb{P}_g$ and $\mathbb{P}_r$. However, this training procedure turns out to be unstable in practice due to mode collapsing of the discriminator and a vanishing gradient, when training with the Jensen-Shannon divergence.

In Wasserstein GAN with Gradient Penalty (WGAN-GP) \cite{gulrajani17}, these problems are alleviated by defining $D$ as a critic, which outputs the Wasserstein-1 distance $W$ \cite{villani09} between $\mathbb{P}_g$ and $\mathbb{P}_r$ instead of a measure of discrimination. Therefore, we refer to $D$ as critic from here on. The WGAN-GP objective function is defined as:

\begin{equation}
    \min_{G} \max_{D \in \mathcal{D}} \mathop{\mathbb{E}}_{\mathbf{x} \sim \mathbb{P}_r} [D(\mathbf{x})] - \mathop{\mathbb{E}}_{\mathbf{\tilde x} \sim \mathbb{P}_g} [D(\mathbf{\tilde x})] ,
\end{equation}

where $\mathcal{D}$ is the set of 1-Lipschitz functions and $\mathbf{\tilde x} = G(\mathbf{z})$. $\mathbf{x} \sim \mathbb{P}_r$ denotes that a sample $\mathbf{x}$ is drawn from the real target distribution $\mathbb{P}_r$. Under an optimal critic, the generator network parameters $\theta_g$ are trained to minimize $W(\mathbb{P}_g, \mathbb{P}_r)$.

\subsection{Loss Function for Magnetic Field Prediction}
In WGAN-GP, a gradient penalty term is added to the standard WGAN loss function:

\begin{equation}
    \begin{aligned}
        L_{WGAN-GP} & = \underbrace{\mathop{\mathbb{E}}_{\mathbf{x} \sim \mathbb{P}_r} [D(\mathbf{x})] - \mathop{\mathbb{E}}_{\mathbf{\tilde x} \sim \mathbb{P}_g} [D(\mathbf{\tilde x})]}_{L_{WGAN}} \\
        & + \underbrace{\lambda_{GP} \mathop{\mathbb{E}}_{\mathbf{\tilde x} \sim \mathbb{P}_{\mathbf{\tilde x}}} [(\Vert \nabla_{\mathbf{\tilde x}} D(\mathbf{\tilde x}) \Vert_2 - 1)^2]}_{L_{GP}} ,
    \end{aligned}
\end{equation}

where $\lambda_{GP}$ is the penalty coefficient and $\mathbb{P}_{\mathbf{\tilde x}}$ is the probability distribution for drawing a sample $\mathbf{\tilde x}$, which is a linear, point-wise combination of a real and a generated sample. The additional loss term ensures that the norm of the gradients of the critic parameters $\theta_d$ is at most 1 on these combined samples from $\mathbb{P}_r$ and $\mathbb{P}_g$, which lets $D$ be an optimal realization of the set of 1-Lipschitz functions. The loss $L_{WGAN-GP}$ is then backpropagated to update the network parameters of the generator and the critic.

Our work is inspired by generative image inpainting from the research area of computer vision, where GANs are trained to inpaint the missing region of a corrupted image $x_0$. Ideally, the generated result $\tilde x$ shall match $x_0$ in all the image pixels available and mimic the ground-truth full image $x$. Hence, a $l1$ loss, $L_{match}$, between the predicted result $\tilde x$ and the given input image $x_0$ and a second $l1$ loss, $L_{mimic}$ between $\tilde x$ and the ground-truth training sample $x$ are formulated:

\begin{equation}
    \begin{aligned}
        L_{match} &  = \| x_0 \odot (\mathbf{1} - \mathbf{m}) - \tilde x \odot (\mathbf{1} - \mathbf{m}) \|_1 ,
        \\
        L_{mimic} & = \| x \odot \mathbf{m} - \tilde x \odot \mathbf{m} \|_1 ,
    \end{aligned}
\end{equation}

where $\mathbf{m}$ is a binary mask with a pixel value of 1 for missing magnetic field values and a value of 0 if field measurements are available. The symbol $\odot$ denotes, here and throughout the paper, the Hadamard product.

For magnetic field prediction, we have additional information of the underlying physics of magnetic fields. We not only want to generate a visual appealing result, we also want the generated magnetic field being constrained by Maxwell's equations. With adding physical loss terms to the loss function, our DL method becomes physics-informed and it can be seen as a regularization for generating magnetic fields. Samples from our target space distribution $\mathbb{P}_r$ are discrete magnetic fields values on a regular grid, $B = \mathbf{x}$. The first physical loss term is Gauss's law for magnetism, which states that:

\begin{equation}
    L_{div} = \nabla \cdot B = 0.
\end{equation}

If we further assume the absence of electric current density $J$ or changing electric field $E$ over time $t$, Ampère's circuital law can be simplified to:

\begin{equation}
    L_{curl} = \nabla \times B = \mu_0 J + \mu_0\epsilon_0 \frac{\delta E}{\delta t} = 0 ,
\end{equation}

where $\mu_0$ is the vacuum permeability and $\epsilon_0$ is the vacuum permittivity.

Our final loss function used during training is formulated as follows:

\begin{equation}
    \begin{aligned}
    L & = \lambda_{WGAN-GP}L_{WGAN-GP} + \lambda_{match}L_{match} \\
    & + \lambda_{mimic}L_{mimic} + \lambda_{div} L_{div} + \lambda_{curl} L_{curl} ,
    \end{aligned}
\end{equation}

where $\lambda_{WGAN-GP}, \lambda_{match}, \lambda_{mimic}, \lambda_{div}$, and $\lambda_{curl}$ are the penalty coefficients for each single loss term and define their relative importance.

\begin{algorithm}
\caption{WGAN-GP for Magnetic Field Prediction}\label{alg:training}
\begin{algorithmic}[1]
\While{$G$ has not converged}
\For{\texttt{5 iterations}}
    \State Sample magnetic fields $\mathbf{B}$ from training data;
    \State Generate random masks $\mathbf{m}$ for $\mathbf{B}$;
    \State Construct input fields $\mathbf{B_{in}} \gets \mathbf{B} \odot (\mathbf{1} - \mathbf{m})$;
    \State Get result $\mathbf{B_{result}} \gets \mathbf{B_{in}} + G(\mathbf{B_{in}}, \mathbf{m}) \odot (\mathbf{m})$
    \State Update $D$ with $L_{WGAN-GP}$
\EndFor
\State Sample magnetic fields $\mathbf{B}$ from training data;
\State Generate random masks $\mathbf{m}$ for $\mathbf{B}$;
\State Update $G$ with $L_{match}$, $L_{mimic}$, $L_{div}$, $L_{curl}$  \\
\qquad \qquad \qquad and $L_{WGAN-GP}$
\EndWhile
\end{algorithmic}
\end{algorithm}

\subsection{Neural Network Architecture}
The DL architecture used for magnetic field prediction is adapted from Yu et al. \cite{yu18} and visualized in Fig. \ref{fig:dl_architecture}. To demonstrate the concept and to make visualization of the results easier, we choose to input a 3-D measured magnetic field in a 2-D rectangular area and output an inter- or extrapolated 3-D magnetic field in this area. These fields are then multiplied with a binary mask $\mathbf{m}$ during training as follows:

\begin{equation}
    B_{in} = B \odot (\mathbf{1} - \mathbf{m}) .
\end{equation}

The two-step generating process is designed in the style of residual learning \cite{he16} and can be described as follows:

\begin{equation}
    \begin{aligned}
        B_{coarse} & = G_{coarse}( B_{in} \odot (\mathbf{1} - \mathbf{m}), \mathbf{m} ) , \\
        B_{fine} & = G_{fine}( B_{coarse} \odot \mathbf{m} + B_{in} \odot (\mathbf{1} - \mathbf{m}), \mathbf{m} ).
    \end{aligned}
\end{equation}

A generator network $G_{coarse}$ generates a coarse prediction by applying a sequence of convolutional layers on $B_{in}$ and the applied mask $\mathbf{m}$. First, the input field is downsampled to a smaller resolution with an increased number of channels, so that same amount of information can be stored with subsequent convolutions being computationally less expensive. Second, several convolutions with differently scaled dilations are performed on the downsampled image to increase the field-of-view of the model and to enable encoding at multiple scales. Finally, the data is upsampled with interpolations to the original size again and results in a coarse prediction $B_{coarse}$.

A second generator network $G_{fine}$ takes $B_{coarse}$ and $B_{in}$ and produces $B_{fine}$ in similar manner as the coarse generator. Parallel to that, a contextual attention branch splits magnetic field is split up in small patches of $3 \times 3$ pixels. The relative importance between these patches and the missing pixels is calculated, which is then used for an improved reconstruction. The idea is here to overcome the locality in the convolutional layers and enhance it with a global information flow from distant magnetic field pixels. The convolution and the attention branch are concatenated before upsampling to the original resolution.

On $B_{fine}$, the loss $L_{match}$, $L_{mimic}$, $L_{div}$, and $L_{curl}$ can be directly calculated. For the adversarial loss $L_{WGAN-GP}$, we need to employ a critic neural network, which tells us the Wasserstein-1 distance between the original and the generated magnetic fields. It has shown beneficial to split the critic in a global critic network $D_{global}$, which evaluates the whole image, and a local critic network $D_{local}$, which determines the quality of the filled-in regions.

\begin{figure}[t]
\centering
\includegraphics[scale=0.55]{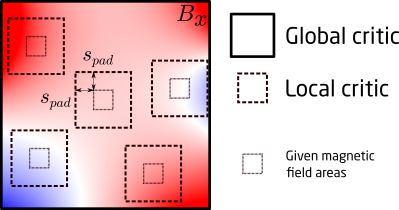}
\caption{Magnetic field regions of the 2-D measurement area, which the two neural networks serving as critic, $D_{global}$ and $D_{local}$, use as input for the outpainting task.}
\label{fig:patch_outpaint}
\end{figure}

In our framework, we extended the setup to work computationally efficient with the outpainting task, which can be seen as an inverted inpainting task. Hereby, only small regions of magnetic field measurements across the 2-D area are given. The missing field values around have to be inter- and extrapolated. Implementing this task in the framework of Yu et al. \cite{yu18} is straightforward. However, special care has to be taken, when creating the local patches for the outpainting task. Instead of naively inverting the mask values and calculating the local patch for nearly the whole image, we define small boxes with with padding size $s_{pad}$ around the given field patch as shown in Fig. \ref{fig:patch_outpaint}.

As the convolutional neural networks used in $G_{coarse}$ and $G_{fine}$ are resolution-independent, the size and shape of $B_{in}$ can vary during inference time. Similarly, the applied missing regions can be arbitrarily chosen by setting the mask pixel values to 1. The complete training procedure is summarized in Algorithm \ref{alg:training}. As usual with GANs, the critic is updated five times before the next update for the generator parameters is performed.

\section{Experiments}

\begin{figure}[t]
\centering
 \begin{subfigure}[b]{0.15\textwidth}
     \centering
     \includegraphics[width=\textwidth]{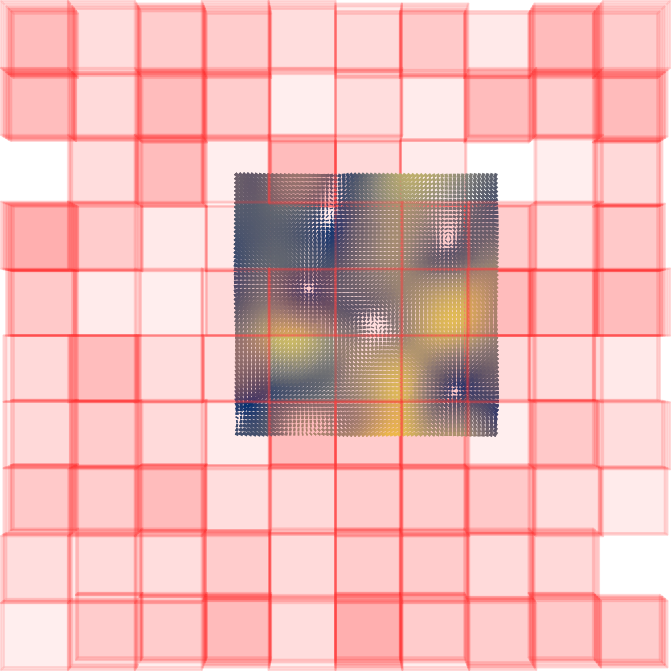}
     \caption{Top view}
     \label{fig:exp_bird}
 \end{subfigure}
 \hfill
 \begin{subfigure}[b]{0.3\textwidth}
     \centering
     \includegraphics[width=\textwidth]{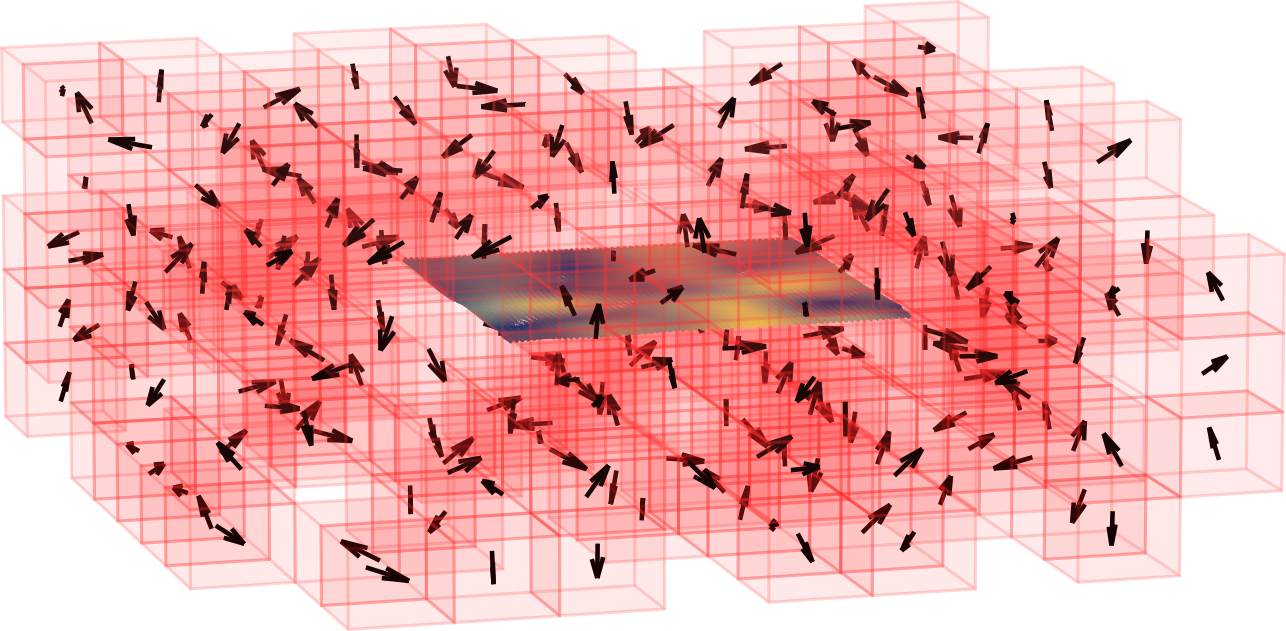}
     \caption{Side view}
     \label{fig:exp_side}
 \end{subfigure}
\caption{Virtual experimental setup. Multiple hard magnets shapes as cubic prisms are placed randomly in a grid of resolution 10 $\times$ 10 $\times$ 5. The magnetic field samples used for training and testing of our novel method are computed in a 2-D area enclosed by this structure. It is assured that no magnetic material can be found in the measurement area.}
\label{fig:virtual_setup}
\end{figure}

To check the performance of our novel method for magnetic field prediction, we introduce a virtual setup, where our open-source micromagnetism and magnetostatic calculation framework MagTense \cite{bjoerk21} is used to place a 3-D construct of hard magnets around a 2-D area and to compute its resulting magnetic field. As shown in Fig. \ref{fig:virtual_setup}, multiple hard magnets are placed randomly with a probability of 50\% in a grid of resolution 10 $\times$ 10 $\times$ 5. Each magnet is shaped as a cubic prism with a fixed side length of 0.1 cm and has a remanent magnetization of 1.2 T, but with the easy axes of the single magnets randomly varying. The different magnetizations and locations of the hard magnets produce a huge variety of magnetic fields in the central area of the grid. The enclosed 2-D rectangular area in the center is left free of magnet material and varying in side length ranging from 0.1 to 0.4 cm between different realizations to include multiple field length scales and a changing number of magnets at the edge of that area. We compute the resulting 3-D magnetic field with a resolution 256 $\times$ 256 pixels and store 20,000 samples of these into a dataset, which is then used to train our neural networks. Additionally, we store a layer of magnetic field calculations above and a layer below to later be able to calculate the divergence and curl of the magnetic flux density with a finite difference method.

\begin{figure}[t]
\centering
 \begin{subfigure}[b]{0.20\textwidth}
     \centering
     \includegraphics[width=\textwidth]{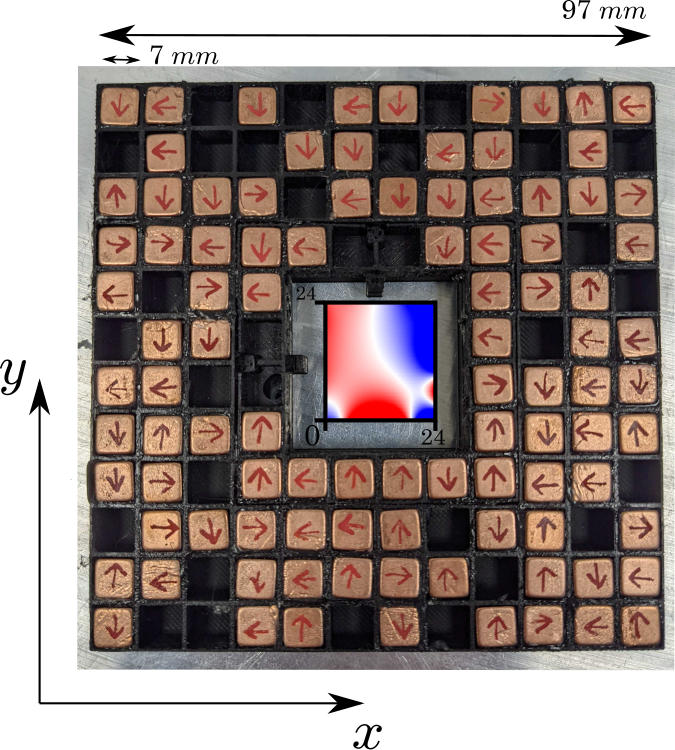}
     \caption{Top view}
     \label{fig:phys_bird}
 \end{subfigure}
 \hfill
 \begin{subfigure}[b]{0.25\textwidth}
     \centering
     \includegraphics[width=\textwidth]{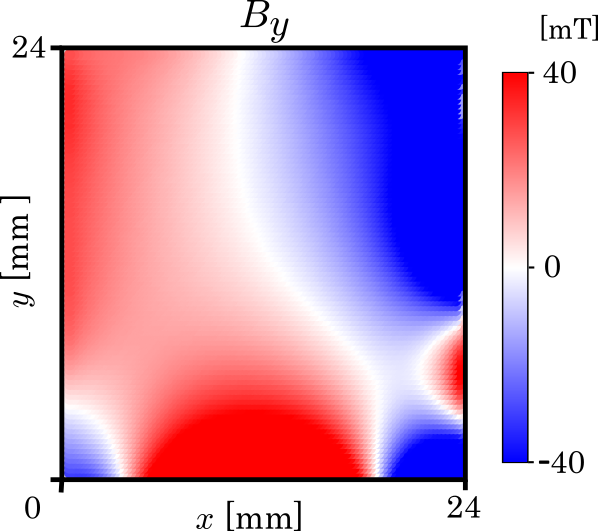}
     \caption{Measured magnetic field}
     \label{fig:phys_side}
 \end{subfigure}
\caption{The physical experimental setup. Fig. \ref{fig:phys_bird} shows the 12 $\times$ 12 holder with the 97 NdFeB magnets. In the enclosed 2-D area, we measure the magnetic field. The $y$-component $B_y$ is visualized in Fig. \ref{fig:phys_side}. For brevity, we omit $B_x$ here and as the easy axes the single hard magnets is in the $xy$-plane, it follows that $B_z = 0$.}
\label{fig:physical_setup}
\end{figure}

Moreover, we built a physical setup with real Neodymium (NdFeB) magnets in our laboratory at Technical University of Denmark and measure the magnetic field with a Hall sensor. We printed a 3-D holder with 12 $\times$ 12 spots and place 97 NdFeB magnets in cubic shape with a side length of 0.7 cm. The hard magnets have a magnitude of 1.29-1.32 T and their easy axis lie in the $xy$-plane. However, production variations lead to small deviations from that plane. In the center of the holder is a hole of size 6 cm $\times$ 6 cm, similar to our virtual experiment. As ground-truth data, we then measure the magnetic field in the enclosed 2-D area. In Fig. \ref{fig:physical_setup} this specific setup is depicted along with the $y$-component of the magnetic field.

For each of these setups, we then perform an inpainting and an outpainting task. For inpainting, a single region in the 2-D measurement area is missing and has to be interpolated. For outpainting, small regions across the measurement area are given and the missing magnetic field values are then inter- and extrapolated with the given information.

We train our method in supervised manner with the created dataset of 20,000 sampled magnetic fields. For inpainting, we extend the vision of the local critic to be 4 pixels larger across the masked area with the idea the generator learns even better to predict magnetic field inserts with a smooth transition across the edge from the given to the predicted area. We define four sub-tasks with varying side lengths of the missing quadratic area. For each of the side lengths of 48, 96, 144, and 192 pixels, we train a separate generator neural network. To generalize better to unseen mask sizes, we further vary the side length of the masks between batches up to 25\% from the side length it is trained for. Each training batch contains of 25 samples for which the loss function is calculated. On the batch loss, the neural network parameters of the generator and the critic are then updated using the Adam optimizer \cite{kingma14} with a learning rate of 1e-4. On an NVIDIA GeForce RTX 3090, this results in an almost full GPU memory usage of its 24 GB. We train each of the setup for at least 300,000 iterations, where the training time differs for different mask sizes. In average, it takes about 2 batches / s, which sums up to a total training time of approximately two days. As a starting point, we take the penalty coefficients directly from Yu et al. \cite{yu18}. The newly introduced hyperparameters $\lambda_{div}$ and $\lambda_{curl}$ are set based on manual inspection of the error magnitude in order to scale $L_{div}$ and $L_{curl}$ to a similar range as the other loss terms. We have visualized the scaled loss terms used for the generator updates in Fig. \ref{fig:in_hyp_com}.  Eventually, the penalty coefficients are defined as follows: $\lambda_{WGAN-GP} = 0.001, \lambda_{GP} = 10, \lambda_{match} = 7.2, \lambda_{mimic} = 3.6, \lambda_{div} = 500, \text{and} \lambda_{curl} = 30,000$.

\begin{figure}[t]
\centering
 \begin{subfigure}[b]{0.24\textwidth}
     \centering
     \includegraphics[width=\textwidth]{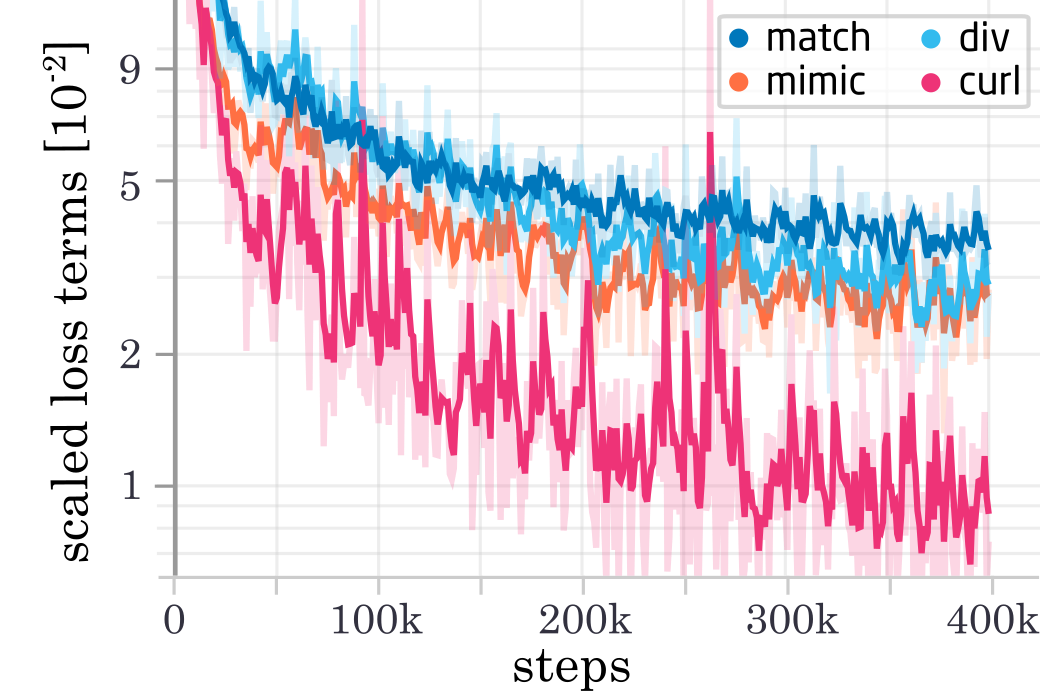}
     \caption{Inpainting - 144 $\times$ 144}
     \label{fig:in_hyp_comp}
 \end{subfigure}
 \hfill
 \begin{subfigure}[b]{0.24\textwidth}
     \centering
     \includegraphics[width=\textwidth]{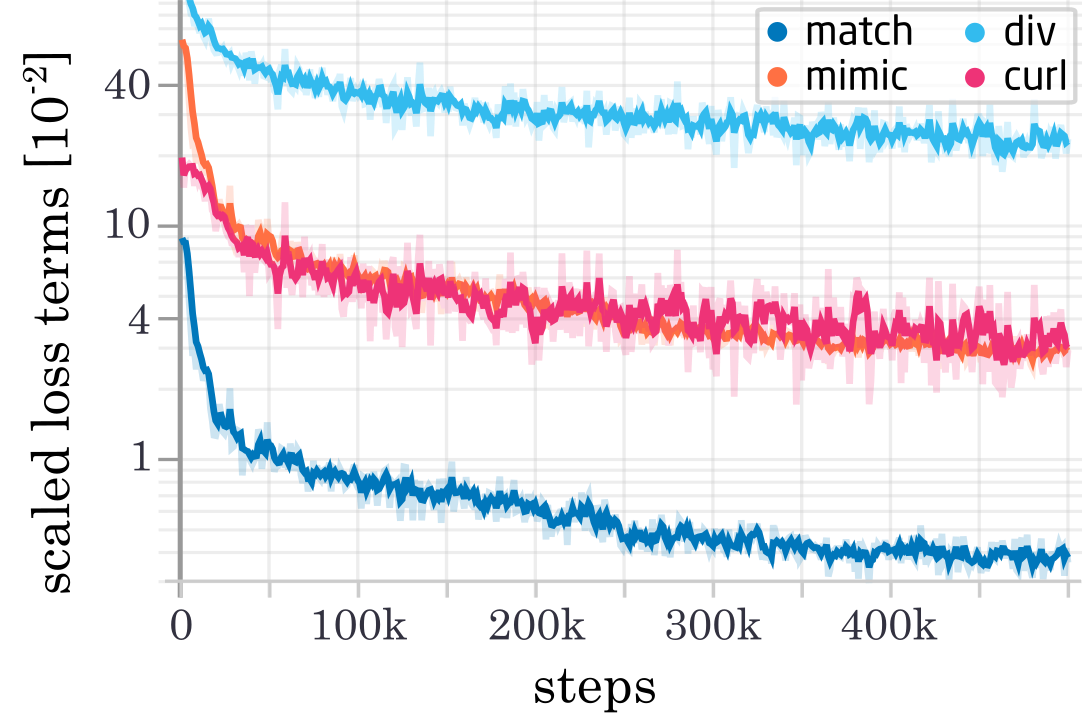}
     \caption{Outpainting - 16 $\times$ 16}
     \label{fig:out_hyp_comp}
 \end{subfigure}
\caption{Overview of the scaled reconstruction and physical loss terms during training. During inpainting, the loss terms are in the same range, whereas in the outpainting task, the divergence loss is an order of magnitude larger than the scaled $L_{mimic}$ and $L_{curl}$. On the other hand, the scaled $L_{match}$ is an order of magnitude smaller compared to these.}
\label{fig:hyp_comp}
\end{figure}

For outpainting, we introduce a setup, where 20 regions of 16 $\times$ 16 magnetic field values across the measurement area are given, and a second setup with only 20 point measurements being available. For each of these setups, we again train a generator neural network similarly to the inpainting task. Now, we use a batch size of 48 samples, which leads to better convergence in this task. Running for 500,000 iterations on two NVIDIA GeForce RTX 3090 in parallel, training time decreases to 1.6 batches / s and sums up to total of around 3.6 days. As the divergence and curl losses become larger in the outpainting task, we adjust $\lambda_{match}$ to 10, $\lambda_{mimic}$ to 2.4, $\lambda_{div}$ to 120, and $\lambda_{curl}$ to 24,000. Fig. \ref{fig:out_hyp_comp} indicates to further decrease $\lambda_{div}$ to obtain more similar ranges of the scaled loss terms in future training runs. A more extensive hyperparameter can additionally reveal an improved set of hyperparameters.

The code, pretrained models, and examples are available at: \url{https://github.com/spollok/magfield-prediction}.

\section{Results}

In the following section, we evaluate our novel method for magnetic field prediction and benchmark its performance with current state-of-the-art methods found in literature. In addition to our method, we solve the tasks with a linear and a cubic spline-based interpolation from SciPy \cite{virtanen20}, biharmonic equations \cite{damelin18}, a Scikit \cite{pedregosa11} implementation of Gaussian processes \cite{williams06} with a radial-basis function kernel, and the adapted WGAN-GP method from Yu et al. \cite{yu18} without the additional physical terms in the loss functions. We skip evaluating Gaussian processes for the inpainting task as its computational complexity scales with $\mathcal{O}(n^3)$, where $n$ is number of given magnetic field measurement, and becomes computationally prohibitive for this task. Moreover, we do not apply linear interpolation for the outpainting task as its implementation does not support extrapolation to magnetic field points outside of the convex hull of the given field measurements.

\subsection{Virtual setup - Inpainting }

\begin{table}[!t]
\renewcommand{\arraystretch}{1.3}
\caption{ MAE [mT] of inpainting task for different mask sizes. The method with the lowest MAE on each sub-task with 250 test samples is marked in bold. Our method is retrained on each of the specific sub-tasks. The subscripts in the two last rows indicate that we evaluate our method trained on one mask size only.}
\label{tab:performance}
\centering
\begin{tabular}{ccccc}
 \hline
   & 48 & 96 & 144 & 192 \\
 \hline
    Linear
        & 3.79 \textpm 3.38
        & 12.43 \textpm 9.80
        & 21.85 \textpm 14.92
        & 30.81 \textpm 18.49 \\
    Spline 
        & \textbf{0.38 \textpm 0.44}
        & \textbf{2.96 \textpm 3.28}
        & 8.46 \textpm 8.54
        & 15.57 \textpm 13.94 \\
   \tiny{Biharmonic \cite{damelin18}}
        & 0.39 \textpm 0.51
        & 3.13 \textpm 3.63
        & 8.60 \textpm 7.94
        & 15.00 \textpm 11.24 \\
    \tiny{WGAN-GP \cite{yu18}}
        & 2.97 \textpm 1.62
        & 3.83 \textpm 2.27
        & 6.69 \textpm 4.38
        & 9.21 \textpm 6.30 \\
    Ours 
        & 2.34 \textpm 1.19
        & 3.51 \textpm 1.85
        & 5.14 \textpm 3.39
        & \textbf{7.51 \textpm 5.29} \\
    $\text{Ours}_{144}$ 
        & 2.53 \textpm 1.18
        & 3.19 \textpm 1.44
        & 5.14 \textpm 3.39
        & 12.62 \textpm 6.61 \\
    $\text{Ours}_{192}$ 
        & 5.82 \textpm 2.05
        & 6.39 \textpm 1.97
        & \textbf{5.09 \textpm 2.50}
        & \textbf{7.51 \textpm 5.29} \\
 \hline
\end{tabular}
\end{table}

\begin{figure*}
     \centering
     \begin{subfigure}[b]{0.13\textwidth}
         \centering
         \includegraphics[width=\textwidth]{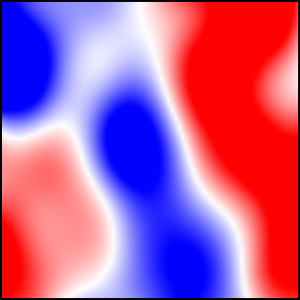}
         \caption{Ground-truth}
         \label{fig:gt}
     \end{subfigure}
     \hfill
     \begin{subfigure}[b]{0.13\textwidth}
         \centering
         \includegraphics[width=\textwidth]{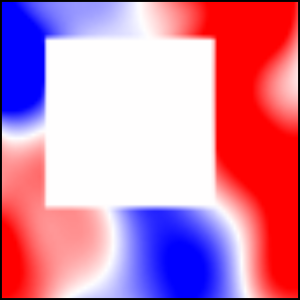}
         \caption{Masked input}
         \label{fig:input}
     \end{subfigure}
     \hfill
     \begin{subfigure}[b]{0.13\textwidth}
         \centering
         \includegraphics[width=\textwidth]{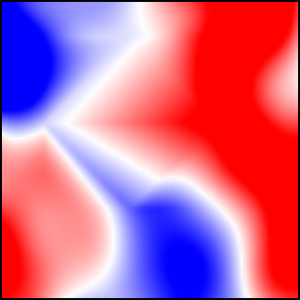}
         \caption{Linear}
        \label{fig:linear}
     \end{subfigure}
     \hfill
     \begin{subfigure}[b]{0.13\textwidth}
         \centering
         \includegraphics[width=\textwidth]{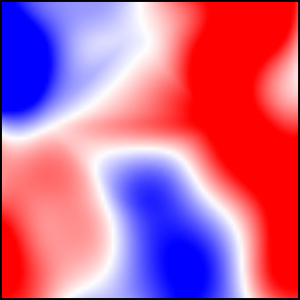}
         \caption{Spline}
        \label{fig:spline}
     \end{subfigure}
     \hfill
     \begin{subfigure}[b]{0.14\textwidth}
         \centering
         \includegraphics[width=0.93\textwidth]{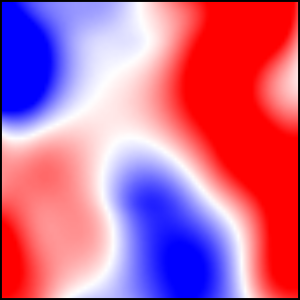}
         \caption{Biharmonic \cite{damelin18}}
         \label{fig:biharmonic}
     \end{subfigure}
     \hfill
     \begin{subfigure}[b]{0.14\textwidth}
         \centering
         \includegraphics[width=0.93\textwidth]{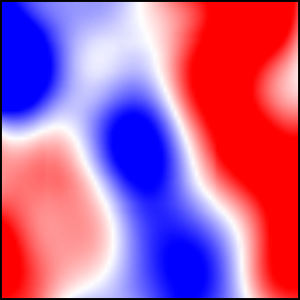}
         \caption{WGAN-GP \cite{yu18}}
         \label{fig:wgan}
     \end{subfigure}
     \hfill
     \begin{subfigure}[b]{0.13\textwidth}
         \centering
         \includegraphics[width=\textwidth]{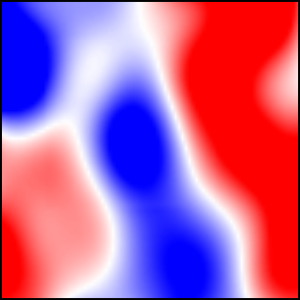}
         \caption{Ours}
         \label{fig:ours}
     \end{subfigure}
        \caption{Qualitative analysis of the inpainting task with a mask of a side length of 144 pixels. Only the trained methods can grasp the curved behaviour of the missing magnetic field. Our method further enhances smoothness and predicts a more correct curvature compared to the WGAN-GP approach.}
        \label{fig:inpaint_144}
\end{figure*}

In Tab. \ref{tab:performance}, we compare our method to four other methods used in literature for magnetic field prediction based on the mean absolute error (MAE) between ground-truth and the predicted magnetic field values in each pixel. Each benchmark is run on 250 test samples, which have not been used during training of the neural networks. Unsurprisingly, linear interpolation shows poor performance the larger the area of missing magnetic field values becomes. An interpolation method based on cubic splines performs the best for small mask sizes with side lengths of 48 and 96 pixels. The biharmonic interpolation performs similarly good, whereas the learning-based WGAN-GP methods, the DL architecture from Yu et al. \cite{yu18} and ours, become valuable with increasing amount of magnetic field values to predict. Our method, the physics-informed version of WGAN-GP, performs best on the masks with side lengths of 144 and 192 pixels. When further comparing the sanity of the magnetic field physics, our method has a divergence loss below 0.25 mT/pixel and a curl loss below 1.25 nT/pixel on all mask sizes, which is substantially lower than the physical losses of the other methods on the two large mask sizes.

To emphasize the advantage of our method on larger masks, we employ a qualitative analysis of the inpainting task with a mask of a side length of 144 pixels. The given magnetic field has a shape of 256 $\times$ 256 pixels for each of the three components $B_x$, $B_y$, and $B_z$. For visualization purposes, we only show the ground-truth field distribution of $B_x$, which is depicted in Fig. \ref{fig:gt}. After masking the input, Fig. \ref{fig:input} serves as the input for the interpolation methods. Only the WGAN-GP methods in Fig. \ref{fig:wgan} and Fig. \ref{fig:ours} grasp the shape in the missing area correctly. The other methods produce sub-optimal results, which can be partly explained with the missing information from the other two magnetic field components, $B_y$ and $B_z$, as these methods are interpolating missing values of one component at a time and hence do not include potentially useful, available information of the magnetic field. On the other hand, the learning-based WGAN-GP approaches act directly on all the three components and process them together.

When further comparing the inference time, i.e., the computation time for predicting the missing magnetic field values in Tab. \ref{tab:inference_time}, the WGAN-GP methods are with 2.48 s competitive with linear interpolation (1.58 s) and cubic splines (1.70 s). The biharmonic equations method needs 43.64 s for one test sample.

\begin{figure}[t]
\centering
\includegraphics[scale=0.45]{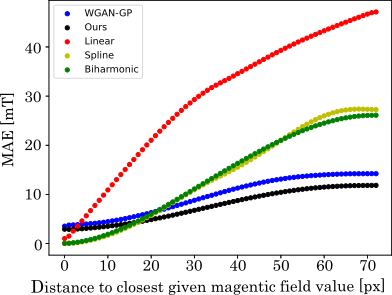}
\caption{Pixel-wise MAE for distance to next given measurement in the input magnetic field on the inpainting task of a masked area of size 144 $\times$ 144. }
\label{fig:px_mae_inpaint}
\end{figure}

For the same task of inpainting a masked area of size 144 $\times$ 144 pixels, we calculate the pixel-wise MAE dependent on the closest given magnetic field measurement. It can be seen in Fig. \ref{fig:px_mae_inpaint} that our method outperforms other interpolation methods the further a magnetic field value to be predicted is away from the measured region. This occurs around a distance of 17 pixels from the mask edge. On smaller distances, there remains a small MAE of around 4 mT, which makes the edge of the predicted field region visually distinguishable from given magnetic fields with low field values. In contrast to the other interpolation methods, the generator $G_{fine}$ generates a full image with a resolution of 256 $\times$ 256 pixels, from which only the masked area is used as prediction for missing field values. Even though the generator is trained with $L_{match}$, it does not succeed to generate magnetic fields that match the given measurements exactly at the edges.

We further investigate how resilient our method is to different mask sizes during inference time compared to the mask size the generator neural network was trained on. We predict the magnetic field values on mask sizes differently from the mask size the generator network was trained on. When using the generator, which was trained on mask sizes with a shape of 144 $\times$ 144 pixels, then, as shown in the second last row of Tab. \ref{tab:performance}, the MAE for smaller mask sizes is similar to the generator specifically trained on that mask size and twice as large on the mask size with a side length of 192 pixels. When using a generator trained on missing regions with a side length of 192 pixels, called $\text{Ours}_{192}$, we can further that the MAE for a mask with a side length of 144 pixels is better than $\text{Ours}_{144}$, which is the generator network trained on that specific mask size, but has a sub-optimal MAE on the smaller mask sizes.

\subsection{Virtual setup - Outpainting}

\begin{figure*}
     \centering
     \begin{subfigure}[b]{0.13\textwidth}
         \centering
         \includegraphics[width=\textwidth]{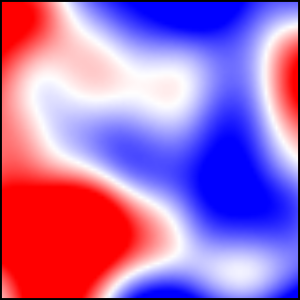}
         \caption{Ground-truth}
         \label{fig:out_gt}
     \end{subfigure}
     \hfill
     \begin{subfigure}[b]{0.13\textwidth}
         \centering
         \includegraphics[width=\textwidth]{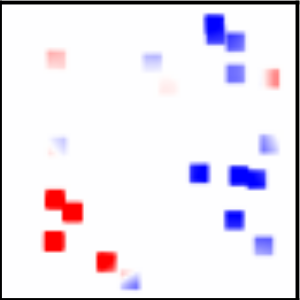}
         \caption{Masked input}
         \label{fig:out_input}
     \end{subfigure}
     \hfill
     \begin{subfigure}[b]{0.13\textwidth}
         \centering
         \includegraphics[width=\textwidth]{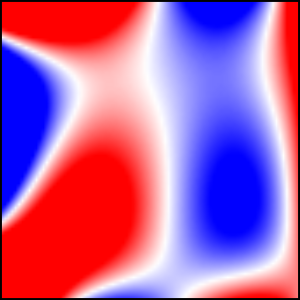}
         \caption{Spline}
        \label{fig:out_spline}
     \end{subfigure}
     \hfill
     \begin{subfigure}[b]{0.14\textwidth}
         \centering
         \includegraphics[width=0.93\textwidth]{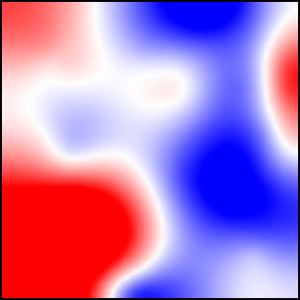}
         \caption{Biharmonic \cite{damelin18}}
         \label{fig:out_biharmonic}
     \end{subfigure}
     \hfill
     \begin{subfigure}[b]{0.13\textwidth}
         \centering
         \includegraphics[width=\textwidth]{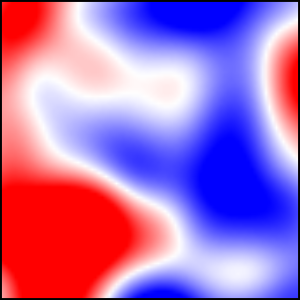}
         \caption{Gaussian}
        \label{fig:out_gaussian}
     \end{subfigure}
     \hfill
     \begin{subfigure}[b]{0.14\textwidth}
         \centering
         \includegraphics[width=0.93\textwidth]{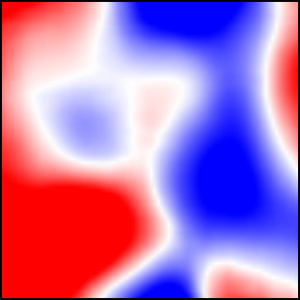}
         \caption{WGAN-GP \cite{yu18}}
         \label{fig:out_wgan}
     \end{subfigure}
     \hfill
     \begin{subfigure}[b]{0.13\textwidth}
         \centering
         \includegraphics[width=\textwidth]{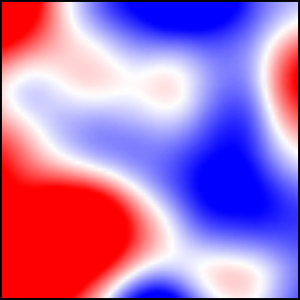}
         \caption{Ours}
         \label{fig:out_ours}
     \end{subfigure}
        \caption{Qualitative analysis of the outpainting task with 20 measurement regions of size 16 $\times$ 16 pixels. Visually, Gaussian processes and our method achieve to reconstruct the ground-truth magnetic field almost perfectly.}
        \label{fig:outpaint_16}
\end{figure*}

In addition to inpainting, we evaluate our method on two outpainting tasks with 100 test samples, respectively. In both task, there are 20 regions of field measurements in each sample available. These regions have the size of 1 $\times$ 1 pixel in the first sub-task and 16 $\times$ 16 pixels in the second one. In Tab. \ref{tab:performance_outpainting}, the reconstruction loss, $L_{mimic}$, and the physical losses, $L_{div}$ and $L_{curl}$ of our method are compared to the losses of four other methods found in literature to perform magnetic field prediction inside and outside of the convex hull of given magnetic field points. Our learning-based, physics-informed method performs best on predicting missing magnetic field values in the setup with 20 point measurements given with a $L_{mimic}$ of 18.60 mT. The biharmonic equations, Gaussian processes, and the WGAN-GP based method achieve comparable results with a reconstruction loss of slightly above 20 mT, whereas the interpolation method based on cubic splines leads to a large representation losses and non-physical predictions, i.e., the divergence and curl of the predicted magnetic field are much greater than 0. For 20 given regions across the measurement area with a side length of 16 pixels, Gaussian processes perform the best with a low $L_{mimic}$ of 3.80 mT and only a small divergence of 0.24 mT / pixel. This is comparable to the error rate in the inpainting task. Our method outperforms the other methods only in the $L_{curl}$ of the magnetic field, while having a twice as high reconstruction error as Gaussian processes for this sub-task. It is important to mention that the Gaussian processes are implemented to process the three components of the magnetic field separately. The good performance of this method is most presumably due to the radial-basis function kernel with a length scale of 10 pixels, which seems to fit well to the smooth nature of magnetic fields and the underlying squared-exponential influence from given magnetic field points to unknown values is able retrieve missing information as can bee seen in Fig. \ref{fig:out_gaussian}. In this example, only our method can visually produce similarly good results as shown in Fig. \ref{fig:out_ours}.
When comparing the inference times of the different methods on the outpainting task with 16 $\times$ 16 pixels, the learning-based WGAN-GP methods have the same computation time of 2.48~s as for the inpainting task. In contrast to the other methods, the size and amount of mask has no influence on the inference time of the DL approach as the computation path from given mask input to prediction of the full magnetic field stays the same. However, as mentioned before, the inference time of Gaussian processes scale with $\mathcal{O}(n^3)$ and evaluates to 233 s for a single image here. The biharmonic equations take 294 s to evaluate, while the spline-based method is the fastest with 0.26 s.

\begin{table}[!t]
\renewcommand{\arraystretch}{1.3}
\caption{Losses for the two outpainting tasks.}
\label{tab:performance_outpainting}
\centering
\begin{tabular}{c|ccc|ccc}
 \hline
     & & 1 $\times$ 1 & & & 16 $\times$ 16 & \\
 \hline
   & $L_{mimic}$ & $L_{div}$ & $L_{curl}$ & $L_{mimic}$ & $L_{div}$ & $L_{curl}$ \\
   &  [mT] & \scriptsize{[mT/px]} & \scriptsize{[$\mu$T/px]} & [mT] & \scriptsize{[mT/px]} & \scriptsize{[$\mu$T/px]} \\
 \hline
    Spline
        & 162 & 7590 & 2228
        & 59.28 & 2.99 & 171 \\
   \tiny{Biharmonic \cite{damelin18}}
        & 24.51 & 1.27 & 5.21
        & 13.90 & 0.79 & 2.77 \\
    Gaussian
        & 20.09 & 0.96 & 4.28
        & \textbf{3.80} & \textbf{0.24} & 1.20 \\
    \tiny{WGAN-GP \cite{yu18}}
        & 22.28 & 1.31 & 7.87
        & 14.05 & 0.86 & 4.57 \\
    Ours 
        & \textbf{18.60} & \textbf{0.87} & \textbf{3.51}
        & 7.85 & 0.39 & \textbf{0.99} \\
 \hline
\end{tabular}
\end{table}

\begin{table}[!t]
\renewcommand{\arraystretch}{1.3}
\caption{Inference time [s] during different sub-tasks.}
\label{tab:inference_time}
\centering
\begin{tabular}{c|ccccc}
 \hline
   Inpainting & Linear & Spline & \tiny{Biharmonic \cite{damelin18}} & \tiny{WGAN-GP \cite{yu18}} & Ours \\
 \hline
   144 $\times$ 144 & \textbf{1.58} & 1.70 & 43.64 & 2.48 & 2.48 \\
 \hline
 \hline
   Outpainting & Gaussian & Spline & \tiny{Biharmonic \cite{damelin18}} & \tiny{WGAN-GP \cite{yu18}} & Ours \\
 \hline
   16 $\times$ 16 & 233.37 & \textbf{0.26} & 294.45 & 2.48 & 2.48 \\
 \hline
\end{tabular}
\end{table}

\subsection{Experimental setup}

\begin{figure}
     \centering
     \begin{subfigure}{0.19\textwidth}
     \begin{subfigure}[b]{0.48\textwidth}
         \centering
         \includegraphics[width=\textwidth]{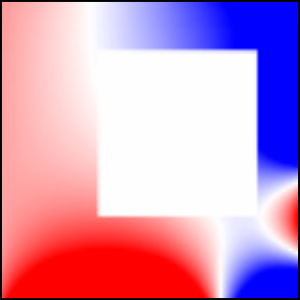}
     \end{subfigure}
     \hfill
     \begin{subfigure}[b]{0.48\textwidth}
         \centering
         \includegraphics[width=\textwidth]{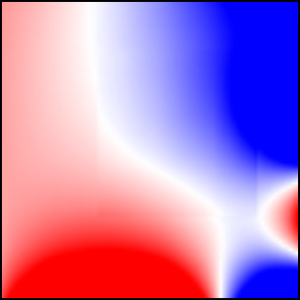}
     \end{subfigure}
     \caption{Inpainting (virtual)}
     \label{fig:phys_qual_in}
     \end{subfigure}
     \hfill
    \begin{subfigure}{0.28\textwidth}
     \begin{subfigure}[b]{0.31\textwidth}
         \centering
         \includegraphics[width=\textwidth]{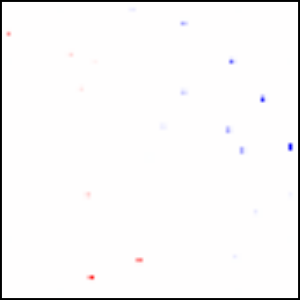}
     \end{subfigure}
     \hfill
     \begin{subfigure}[b]{0.31\textwidth}
         \centering
         \includegraphics[width=\textwidth]{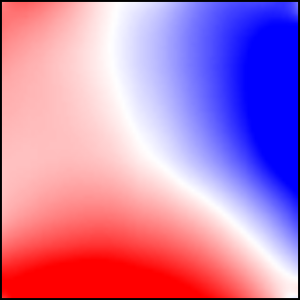}
     \end{subfigure}
     \hfill
     \begin{subfigure}[b]{0.31\textwidth}
         \centering
         \includegraphics[width=\textwidth]{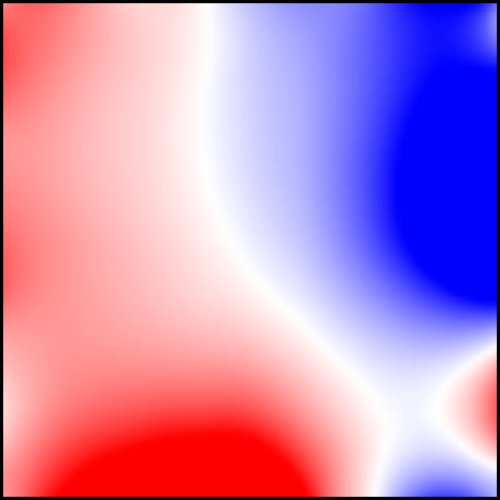}
     \end{subfigure}
     \caption{Outpainting (virtual, experimental)}
     \label{fig:phys_qual_out}
     \end{subfigure}
    \caption{Qualitative analysis on magnetic field prediction in the experimental 2-D setup with a generator network that was trained on the virtual setup. For outpainting, the prediction of a generator retrained on the experimental setup in the same masked input is shown as well. For each of the tasks, the masked input and the predicted magnetic field of $B_y$ are visualized.}
    \label{fig:phys_qualitative}
\end{figure}

To further validate the performance and generalizability of our approach, we use the trained generator of our learning-based, physics-informed WGAN-GP approach, which was trained on the 3-D virtual experimental setup from Fig. \ref{fig:virtual_setup}, to make predictions in the 2-D physical setup as shown in Fig. \ref{fig:physical_setup}. Therefore, we have measured 8,342 magnetic field points in the enclosed 2-D area of size 24 $\times$ 24 mm with a Hall sensor. Hereby, we obtain a resolution of 96 $\times$ 86 inside that area. We again perform an inpainting task with a mask size of 48~$\times$~48 pixels and an outpainting task with 16 regions of size 8~$\times$~8 pixels given. The qualitative results are shown in Fig. \ref{fig:phys_qualitative} with the generator trained on the virtual setup with a mask size with a side length of 144 pixels for inpainting and with the generator network trained on the outpainting task with mask size of 1 $\times$ 1 pixel. Additionally, we show the prediction of a generator, which is retrained on magnetic fields resulting from a virtual, rebuilt setup that is similar to the experimental setup in Fig. \ref{fig:phys_bird}. In the new dataset, the 128 empty spots are randomly filled with hard magnets being only magnetized in the $xy$-plane, i.e., the z-component of their magnetization is to 0. The inpainting results seem to agree well with the original magnetic field. The predictions of the outpainting task though show a clear visual disagreement in the lower right part of the enclosed 2-D area. However, the field predictions in that area substantially improved, when retraining a generator network on magnetic fields similar to the test field.

\section{Discussion}

\begin{figure}[t!]
    \centering
    \begin{subfigure}[t]{0.235\textwidth}
        \centering
        \includegraphics[width=\textwidth]{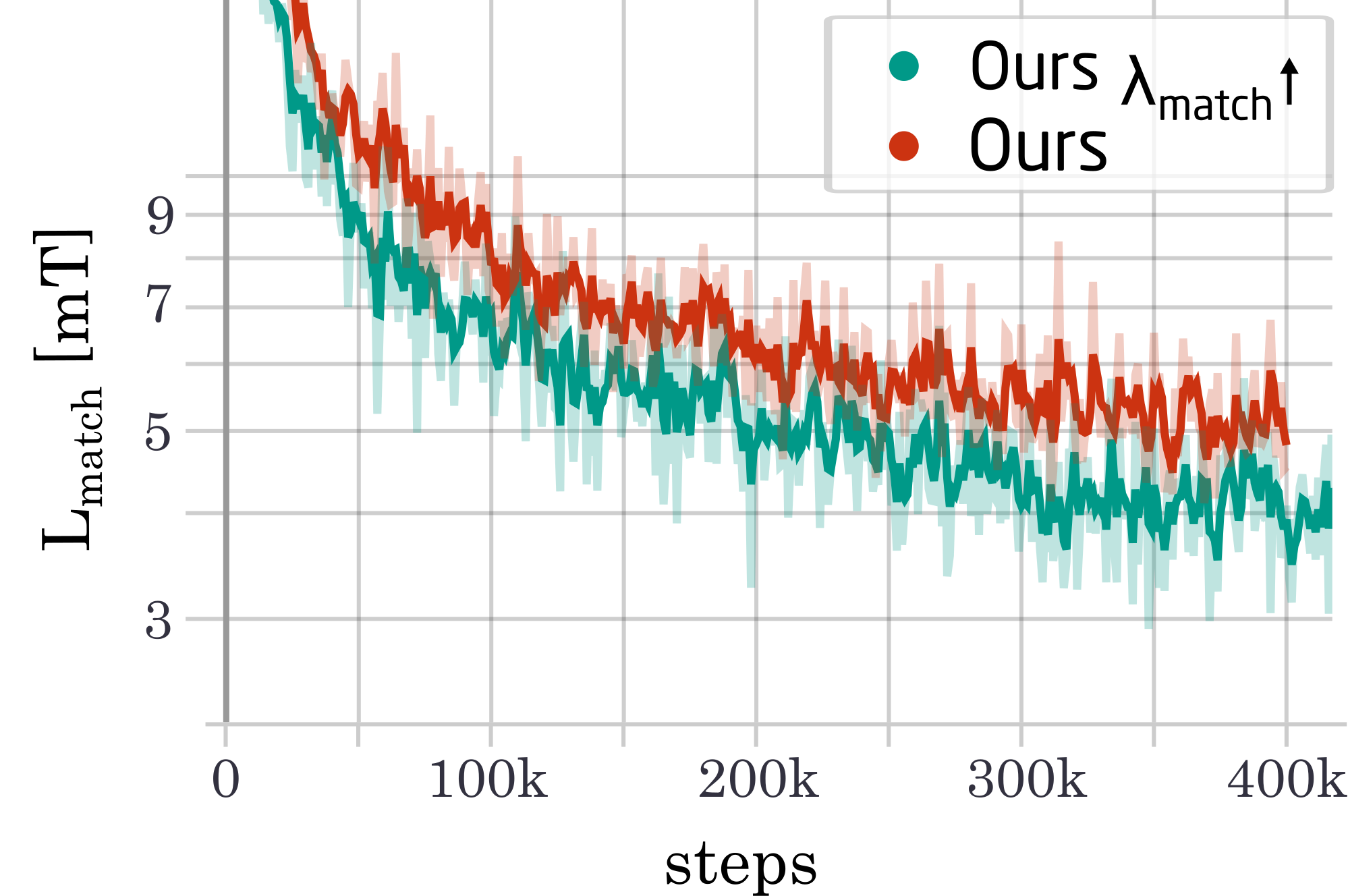}
        \caption{$L_{match}$}
        \label{fig:ae_hyp_tuning}
    \end{subfigure}
    \hfill
    \begin{subfigure}[t]{0.235\textwidth}
        \centering
        \includegraphics[width=\textwidth]{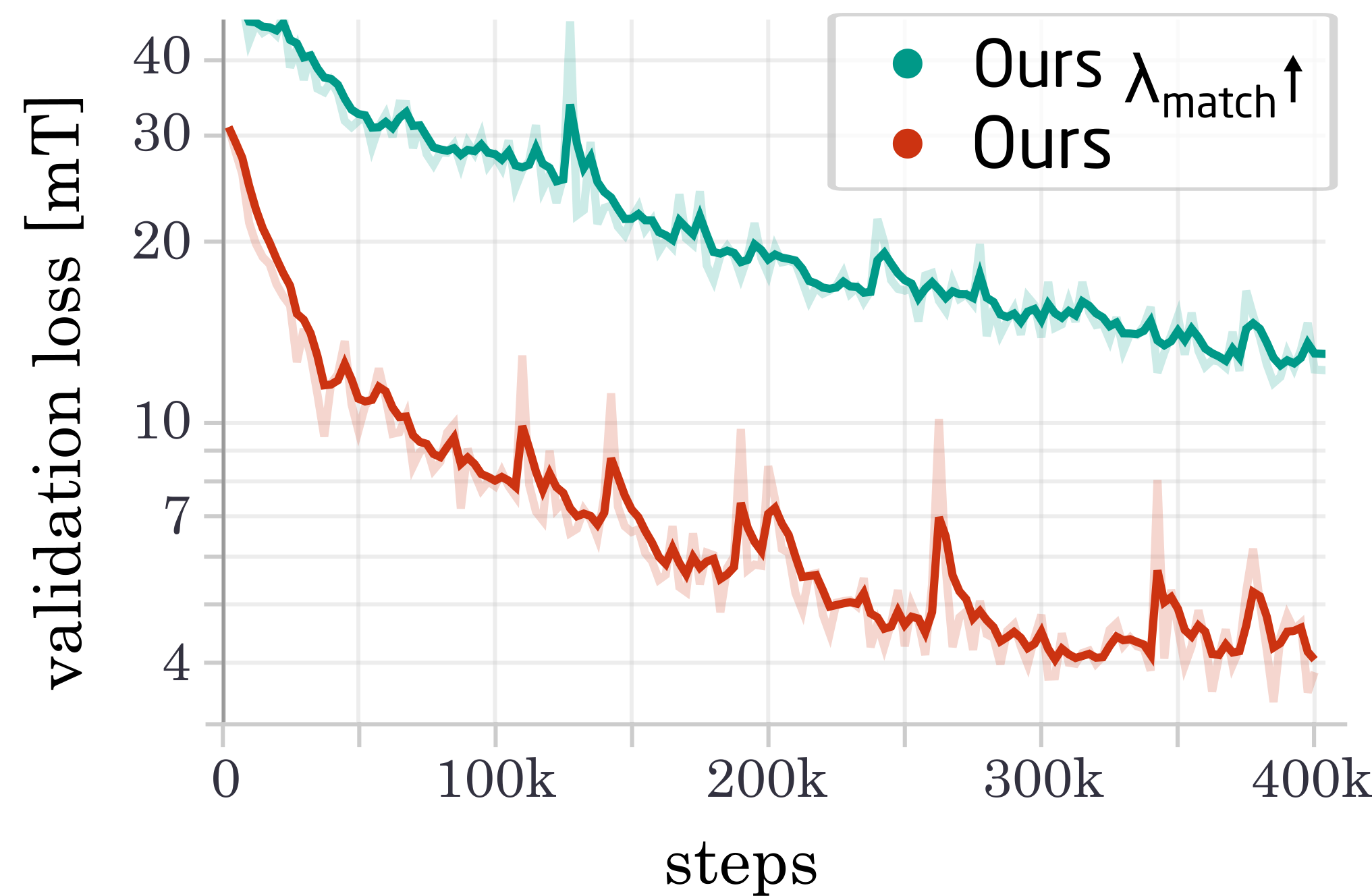}
        \caption{Validation loss}
        \label{fig:val_loss_hyp_tuning}
    \end{subfigure}
    \caption{Comparison of $L_{match}$ and validation loss during training for the inpainting task with different values for $\lambda_{match}$. The red curve shows the hyperparameter set chosen throughout the study with $\lambda_{match} = 7.2$. The green curve uses a $\lambda_{match} = 1000$ to enhance to importance of this error with the intention to decrease the mismatch of generator predictions on locations with known magnetic field measurements.}
\end{figure}

The DL approach is working well and better than other methods found in literature, when the area of missing magnetic field measurements is becoming large. However, our approach has flaws when the value to predict is close to the region of given field points. As shown in Fig. \ref{fig:px_mae_inpaint}, the MAE for standard interpolation methods becomes very low as the given closest points are used as starting points for the specific interpolation technique. On the other hand, the WGAN-GP approaches take the given field points as input to the DL architecture and over several convolutions outputs a prediction, which is only indirectly coupled to the available measurements and the reconstruction of these given magnetic field points is solely controlled by $L_{match}$, which the generator tries to minimize over several updates of its network parameters. An obvious first idea to tackle this issue would be to increase $\lambda_{match}$ and hence the importance of the loss. During parameter updates more focus will be put on matching the given original points. We trained a new generator network with an updated hyperparameter set. As visualized in Fig. \ref{fig:ae_hyp_tuning}, the green curve has indeed an improved, lower $L_{match}$, indicating that the prediction matches the masked input in the given areas better. But when looking at the overall validation loss in Fig. \ref{fig:val_loss_hyp_tuning}, one can see that the calculated MAE on samples unseen during training is larger throughout the training. This can be explained with an increase of the other losses $L_{div}$ and $L_{curl}$, which leads to a less physical model and worse performance on the magnetic field prediction in unknown areas. 
In general, a more elaborate hyperparameter research could substantially enhance the performance of our method. Due to limited available calculation time, we performed all our experiments with these set of hyperparameters without further tuning. 
Another approach to alleviate this behaviour could be a post-processing method to smooth values at the edges of the final result or to combine it with a spline-based method close to given field points. For instance, one can predict values close to given measurements with cubic spline-based interpolation and at about a distance of 17 pixels from the next given measurements the magnetic field predictions from our method can be more and more taken into account. 

\begin{figure}[t!]
    \centering
    \begin{subfigure}[t]{0.235\textwidth}
        \centering
        \includegraphics[width=\textwidth]{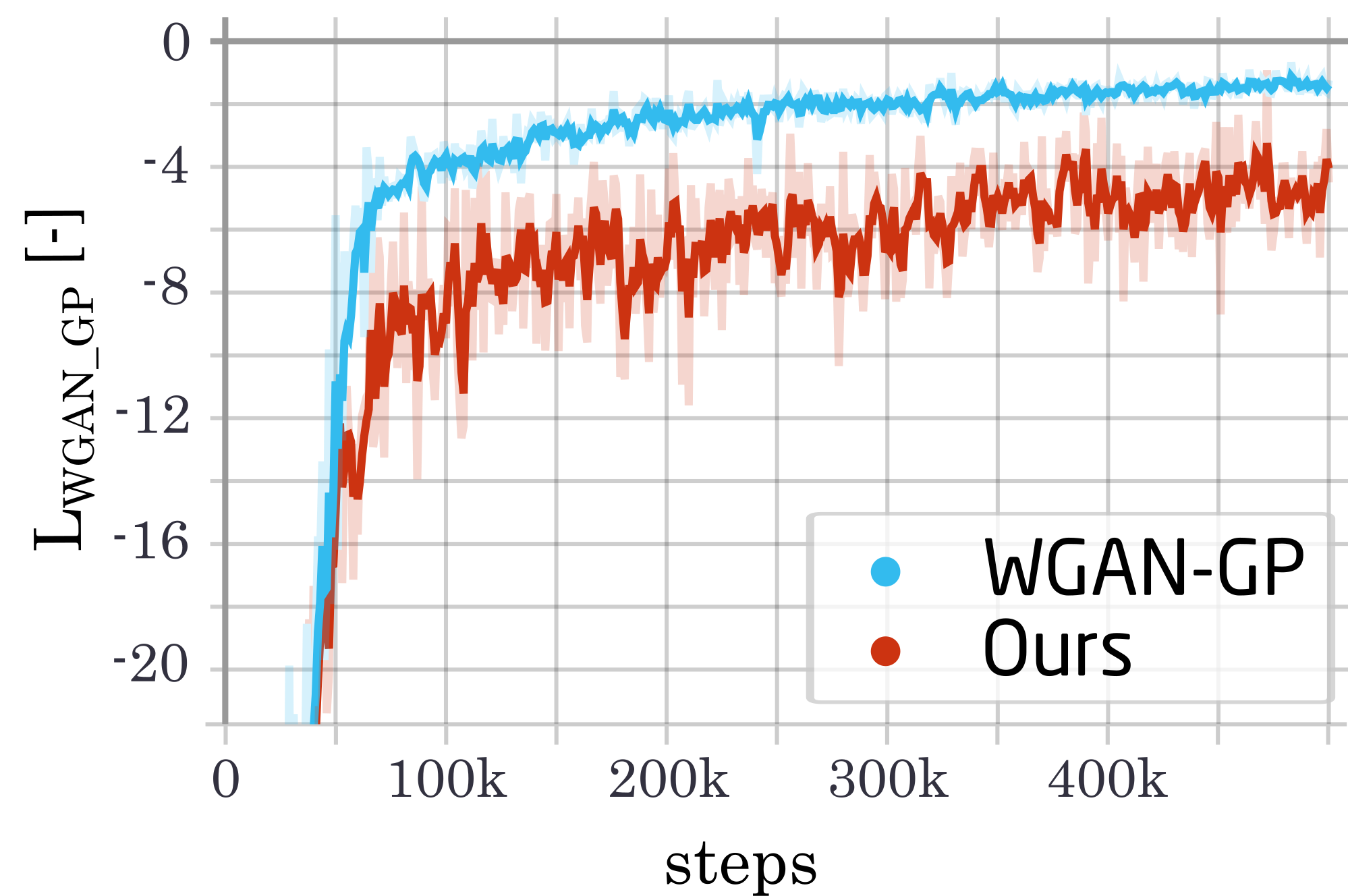}
        \caption{$L_{WGAN\_GP}$}
    \end{subfigure}%
    ~ 
    \begin{subfigure}[t]{0.235\textwidth}
        \centering
        \includegraphics[width=\textwidth]{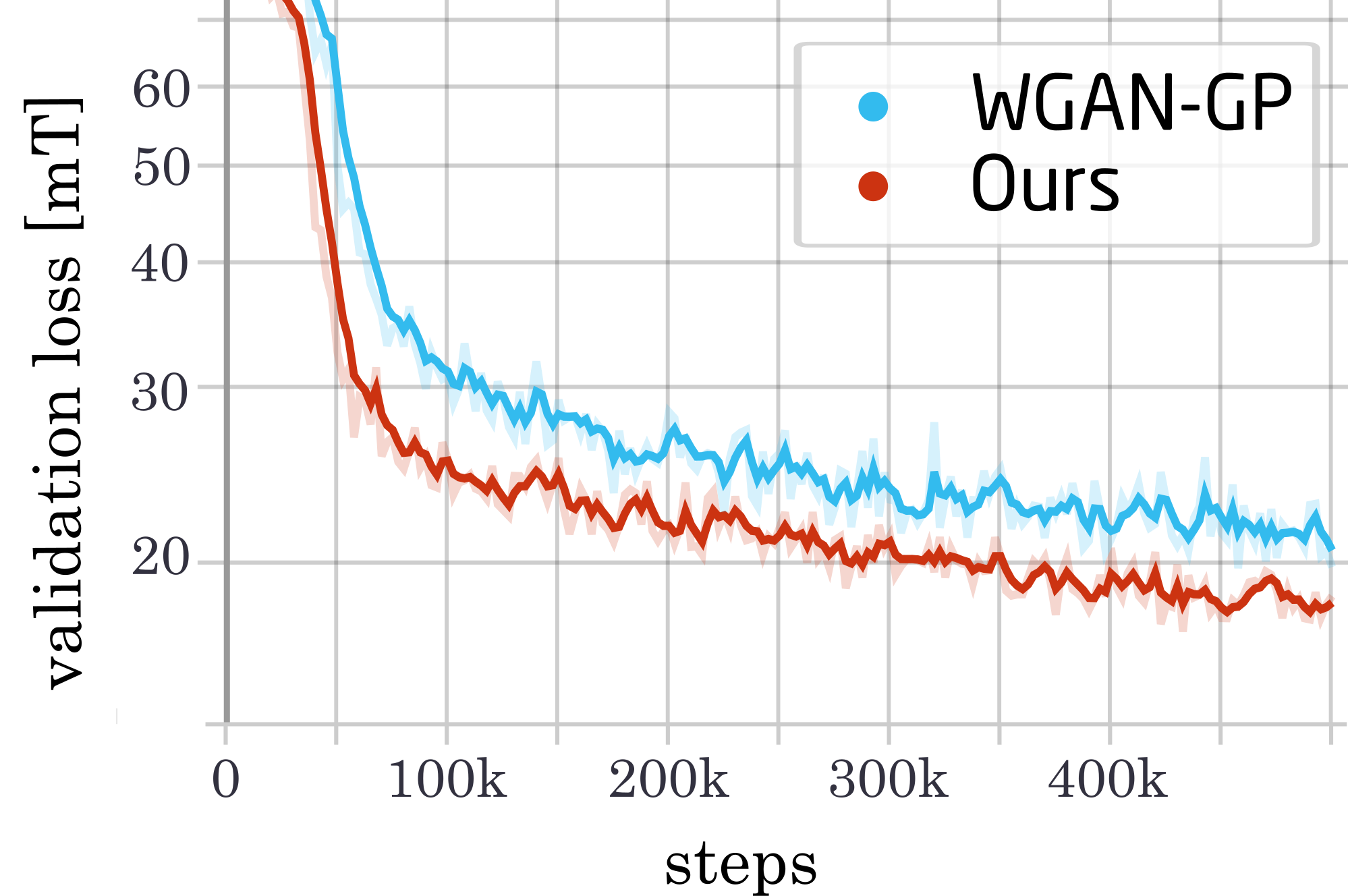}
        \caption{Validation loss}
    \end{subfigure}
    \caption{Training curves for the outpainting task. The blue curve shows the progress of $L_{WGAN\_GP}$ and the validation loss with the standard WGAN-GP method. The red curve is the evolution during training, when enabling physical losses with our method.}
    \label{fig:out_phys_loss_comparison}
\end{figure}

Another interesting point to discuss is the improvement of our physics-informed method compared to the standard WGAN-GP approach. Especially, on the outpainting task our method leads to a substantially lower MAE error. With Fig.~\ref{fig:out_phys_loss_comparison}, we want to emphasize the importance of introducing the physical losses into the DL architecture. The introduction of the physical losses influences the updates of the generator network parameters to be more general, which in return makes it easier for the critic to differentiate between real and generated samples and $L_{WGAN\_GP}$ stays lower throughout the training. We recall that the critic networks tries to maximize the Wasserstein-1 distance between the real and the generated magnetic field distribution. Hence, $L_{WGAN\_GP} = 0$ means that the critic cannot distinguish between real and generated samples anymore. Consequently, the generator network has less incentives to improve its generating process of magnetic field predictions. Eventually, this will then result in a higher $L_{val}$.

When looking at Fig. \ref{fig:phys_qual_out} one can see that the trained generator is performing well on large parts of $B_y$ to predict, but has difficulties to anticipate the fast switching magnetic field in the lower right part of the measurement area. These emerge from the adjacent hard magents at the border of the enclosed 2-D area as visualized in Fig. \ref{fig:phys_bird}. In comparison to the virtual 3-D setup, the gap is smaller here and hence the magnetic field produced in the lower right area is not part of the magnetic field distribution the generator was trained on. A remedy for that could be either to include such magnetic field data in the training data to make the generator predictions even more general or to completely retrain on the physical 2-D setup.
A further feature to implement could be an additional input parameter to signalize a specific condition, e.g., the number magnets at the border the enclosed 2-D area or other geometrical implications such as the gap between magnets and measurements.

\section{Conclusion}
With our novel method, we are able to perform magnetic field prediction better than current state-of-the-art methods on inpainting tasks, where large parts of the magnetic field measurement relative to the overall measurement area is missing. Moreover, our physics-informed, learning-based method produces the best results when comparing it to other state-of-the-methods on an outpainting task with only point measurements (1 $\times$ 1 pixel) given. When regions (16 $\times$ 16 pixels) of measurements are given, then Gaussian processes outperform our method, however, with the inference time of magnetic field prediction being two order of magnitude higher. In some time critical applications as the simultaneous mapping and localization performed in robotics mentioned in the introduction, our model could in this task serve as a trade-off between accuracy and computational time.

\section*{Acknowledgements}
We want to thank our colleague Sina Jafarzadeh for performing the physical measurements in our laboratory manually with a Hall sensor. Every setup variation or change in resolution entailed another time-consuming measurement process.

\end{document}